%% file: main_arxiv.tex
\documentclass[10pt,journal,compsoc]{IEEEtran}

\usepackage[nocompress]{cite}
\usepackage[pdftex]{graphicx}
\usepackage{amsmath}
\usepackage{algorithmic}
\usepackage{array}
\usepackage{fixltx2e}
\usepackage{url}
\usepackage{epsfig}
\usepackage{graphicx}
\usepackage{comment}
\usepackage{amsmath}
\usepackage{amssymb}
\usepackage[linesnumbered, ruled,vlined]{algorithm2e}
\usepackage{caption,booktabs,array}
\usepackage{multirow}
\usepackage{tikz}
\usepackage{blkarray}
\usepackage{xcolor}
\usepackage{wrapfig}

\def\balpha{{\boldsymbol \alpha}}
\def\bbeta{{\boldsymbol \beta}}

\usepackage[utf8]{inputenc}
\def\dd{{\boldsymbol d}}
\def\rr{{\boldsymbol r}}
\def\xx{{\boldsymbol x}}

\def\yy{{\boldsymbol y}}

\def\CC{{\boldsymbol C}}

\def\HH{{\boldsymbol H}}
\def\UU{{\boldsymbol U}}
\def\TT{{\boldsymbol T}}

\def\balpha{{\boldsymbol \alpha}}
\def\bbeta{{\boldsymbol \beta}}

\input{math_commands.tex}

\usepackage{url}

\newcommand{\kf}[1]{{\color{black} #1}}

\usepackage[normalem]{ulem}

\newtheorem{proposition}{Proposition}

\ifCLASSOPTIONcompsoc
  \usepackage[nocompress]{cite}
\else
  \usepackage{cite}
\fi
\ifCLASSINFOpdf
\else
\fi
\begin{document}
%
%
%
%
%

\title{{Wasserstein Adversarial Regularization \\for learning with label noise}}

\author{Kilian~Fatras$^\dagger$,Bharath Bhushan~Damodaran$^\dagger$,
         Sylvain~Lobry, R\'emi~Flamary,  Devis~Tuia,~\IEEEmembership{Senior Member,~IEEE,}
        and~Nicolas~Courty
        
\thanks{$\dagger$ Authors contributed equally}
\thanks{K. Fatras and N. Courty are with the University Bretagne Sud, CNRS, INRIA, IRISA, UMR 6074, France, e-mail: kilian.fatras@irisa.fr}
\thanks{B.B. Damodaran is with InterDigital R\&D, Rennes, France. This work was done during the post-doctoral position at University Bretagne Sud, CNRS, IRISA, UMR 6074, France.}%
\thanks{S. Lobry is with LIPADE, Université de Paris, Paris, France. This work was done during the post-doctoral position at Laboratory of Geo-Information Science and Remote Sensing, Wageningen University, The Netherlands}%
\thanks{R. Flamary is with CMAP, École Polytechnique.}%
\thanks{D. Tuia is with Ecole Polytechnique Fédérale de Lausanne, Sion, Switzerland.}
\thanks{Manuscript received XXXXXXX.}}


%
%

\markboth{Journal of \LaTeX\ Class Files,~Vol.~14, No.~8, August~2015}%
{Shell \MakeLowercase{\textit{et al.}}: Bare Demo of IEEEtran.cls for Computer Society Journals}
%



\IEEEtitleabstractindextext{%
\begin{abstract}

 Noisy labels often occur in vision  datasets, especially when they are obtained from crowdsourcing or Web scraping. We propose a
   new regularization method, which enables learning  robust classifiers in presence of noisy data. To achieve this goal, we propose a new adversarial regularization {scheme} based on the Wasserstein distance. Using this distance allows taking into account specific relations
   between classes by leveraging the geometric properties of the labels space. 
   {Our Wasserstein Adversarial Regularization (WAR) encodes a selective regularization, which promotes smoothness of the classifier between some classes,  while preserving sufficient complexity of the decision boundary between others. We first discuss how and why adversarial regularization can be used in the context of 
   noise and then show the effectiveness of our method on five datasets corrupted with noisy labels: in both benchmarks and real datasets, WAR outperforms the state-of-the-art competitors.} 
\end{abstract}


}

\maketitle

\IEEEdisplaynontitleabstractindextext

%
\IEEEpeerreviewmaketitle

\IEEEpeerreviewmaketitle

\input{sub_part/intro.tex}
\input{sub_part/WAT.tex}

\input{sub_part/experiments.tex}

\input{sub_part/conclusion.tex}
\section*{Acknowledgment}
This work was partially funded through projects OATMIL ANR-17-CE23-0012 and 3IA Cote d’Azur Investments ANR-19-P3IA-0002  of French National Research Agency (ANR). This research was supported by 3rd Programme d’Investissements d’Avenir [ANR-18-EUR-0006-02]. This action benefited from the support of the Chair "Challenging Technology for Responsible Energy" led by l’X – Ecole polytechnique and the Fondation de l’Ecole polytechnique, sponsored by TOTAL

\bibliographystyle{IEEEtran}
\bibliography{IEEEabrv,egbib}

\appendix

\input{sub_part/appendix.tex}

\end{document}

%% file: math_commands.tex
\usepackage{amsmath,amsfonts,bm}

\def\eqref#1{equation~\ref{#1}}

\def\1{\bm{1}}

\def\rr{{\textnormal{r}}}

\DeclareMathAlphabet{\mathsfit}{\encodingdefault}{\sfdefault}{m}{sl}
\SetMathAlphabet{\mathsfit}{bold}{\encodingdefault}{\sfdefault}{bx}{n}

\DeclareMathOperator*{\argmin}{arg\,min}

%% file: sub_part/intro.tex

\section{Introduction}
\IEEEPARstart{D}{eep}
neural networks 
require large amount of accurately annotated training samples to achieve good generalization performances. Unfortunately, annotating large datasets is a challenging and costly task, which is practically impossible to do perfectly for every task at hand.
It is then most likely that datasets will contain incorrectly labeled data, which induces noise in those datasets and can hamper learning. This problem is often referred to as \emph{learning with label noise} or \emph{noisy labels}. 
The probability {of facing} this problem increases when the dataset contains several fine grained classes that are difficult to distinguish \cite{Schroff2011,Krause16,Dub18}. 
 As pointed
out in \cite{Zhang_2017}, deep convolutional neural networks have huge
memorization abilities and can learn very complex functions. That is why training with noisy data labels can lead to poor generalization \cite{Arp17,Wang2018,Choi2018}. Classifiers like SVM also suffer from noisy label \cite{biggio112011}.
In this paper we propose a method tackling the problem of overfitting on noisy labels, and this without access to a clean validation dataset.

This problem has been considered in recent literature, mainly in three ways. First are \emph{data cleaning methods}: \cite{Brodley1999} uses a set of learning algorithms to create classifiers that serve as noise filters for the training data. \cite{Vahdat17,Xiao2015,Li2017} learn relations between noisy and clean labels before estimating new labels for training. In \cite{Lee2017cleannet}, few human verified {labels} were necessary to detect noisy labels and adapt learning. Authors in \cite{Natarajan_2013} use a simple weighted surrogate loss to mitigate the influence of noisy labels. In \cite{MentorNet2018,Ren18}, the methods rely either on a curriculum or on meta-gradient updates to re-weight training sets and downweight samples with noisy labels. Second are \emph{Transition probability-based methods}: \cite{Liu_2014, menon_2015, SukhbaatarF14,Patrini_2017_CVPR,Hendrycks2018} estimate a probability for each label to be flipped to another class and use these estimations to build a noise transition matrix. In \cite{SukhbaatarF14}, the authors add an extra linear layer to the softmax in order to learn the noise transition matrix itself, while \cite{Hendrycks2018} uses a small set of clean validation data to estimate it. \cite{Patrini_2017_CVPR} proposes a forward/backward loss correction method, which exploits the noise transition matrix to correct the loss function itself. Third are \emph{regularization-based methods:}
{ In \cite{Reed2015,D2L18}, the authors use a mixture between the noisy labels and network predictions. In \cite{Daiki2018,Kun19}, the regularization is achieved by alternatively optimizing the network parameters and estimating the true labels while the authors of \cite{Coteaching_2018,Yu19,Hongxin20}
propose peer networks feedbacking each other about predictions for the noisy labels. \cite{Song19} proposes to replace noisy labels in the mini-batch by the consistent network predictions during training, while \cite{Chen19}
proposes noisy cross-validation to identify samples that have correct labels. In \cite{Wang19,Zhang2018,Ghosh2017}, robust loss functions are proposed to overcome limitations of cross entropy loss function.}

\begin{figure}[!t]
 \centering
 \includegraphics[width=.49\textwidth]{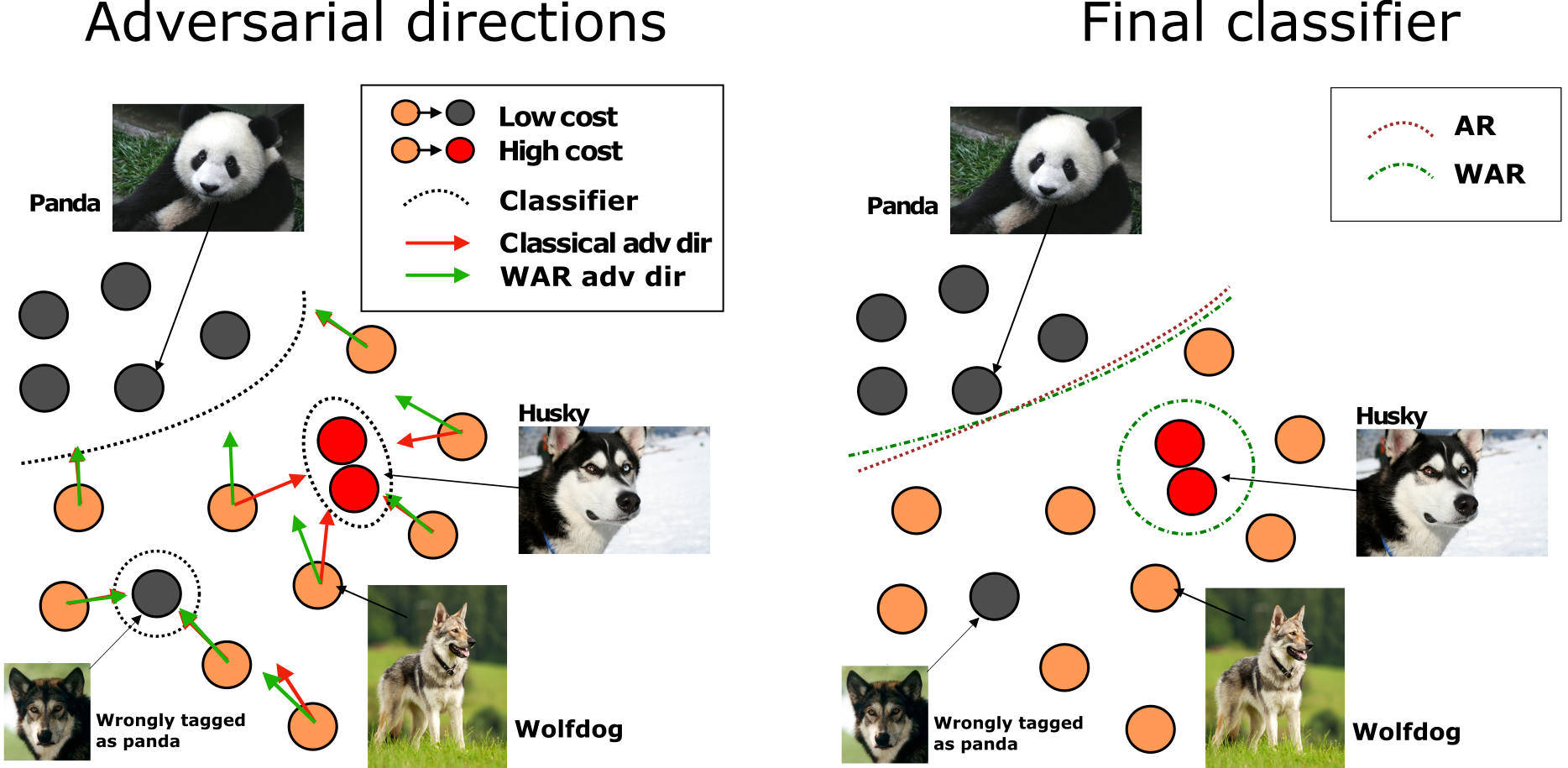}
 \caption{AR vs. WAR. Given a number of samples, both methods regularize along adversarial directions (arrows in the left panel), leading to updated decision functions (right panel).  While both regularizations prevent the classifier to overfit on the noisy labelled sample, AR also tends oversmooth between similar classes ({\em wolfdog} and {\em husky}), while WAR preserves them by changing the adversarial direction.}
 \label{fig:fig00}
\end{figure}

In contrast to those works, we propose to regularize model predictions  according to similarity between classes: by doing so, we allow for learning complex class boundaries between similarly looking classes (i.e., we take different labels in high consideration), while having simpler class boundaries between classes showing low similarity (i.e. we mitigate the effect of label noise).
To do so, we use the adversarial {regularization (AR)} framework ~\cite{Goodfellow2015,VATMiayto2016} on the noisy label problem.
 We extend AR to reduce the discrepancy between the
 prediction of a true input sample and the one obtained by a near-by
 adversarial sample. As a discrepancy, we use a loss {based on the
 Wasserstein distance} computed with respect to a ground cost {encoding}
 class similarities. This ground cost provides 
 the flexibility to regularize with different strengths pairs of classes{. This strength can depend on semantic relations, classes similarities, or prior knowledge (e.g. on annotators' mistakes).}
  This way, the classifier can discriminate non-similar
 objects {robustly} under the presence of noise and
 {class overlap}. {We name our proposed method \emph{Wasserstein Adversarial Regularization} (WAR).}
{WAR} allows incorporating specific knowledge about the potential degree of mixing of classes
through a ground cost that can be designed {\em w.r.t.} the problem at hand. Nevertheless, this knowledge might be unknown or difficult to craft. The idea of considering class or data similarities can be found in many machine learning applications. For instance, several work on clustering relied on creating a graph between data in order to estimate data similarities \cite{Cristianini_2001, Li_noisy_spec_clustering, Yuan2008}. In a remote sensing application, \cite{Volpi_2015_CVPR_Workshops} used a structured SVM to learn an informative set of appearance descriptors jointly with local class interactions.
{In this paper, and among others, we use distances between word embeddings of the class names to derive a semantic ground cost. Experiments in five datasets (Fashion-MNIST, CIFAR10, CIFAR100 and real life examples on clothing classification and a remote sensing semantic segmentation dataset) under label noise conditions show that WAR outperforms the state-of-the-art in providing robust class predictions.}
\newline

%% file: sub_part/WAT.tex
\section{{Wasserstein Adversarial Regularization for label noise}}\label{sec:war}
Given a set of labeled data $\{\xx_i,\yy_i\}_{i=1,\hdots N}$, we are interested in learning a classifier  $p_{\theta}$ defined by a set of parameters $\theta$. Data $\xx_i$ are usually elements of $\mathbb{R}^d$, while $\yy_i \in \mathcal{C}$ are one-hot vectors encoding the belonging to one of $C$ classes.  
The empirical risk minimization principle is used to learn  $p_{\theta}$. Given a loss function $L$, the optimal set of parameters for the classifier is given by $\argmin_\theta \sum_{i=1}^N L(\xx_i,\yy_i,p_\theta)$.  Without loss of generality, we will consider that $L$ is the cross-entropy loss: \kf{ $L_{\text{CE}}(\xx_i,\yy_i,p_\theta) = - \sum_{c=1}^{C} \yy_i^{(c)} \log  p_\theta(\xx_i)^{(c)}$. }

The considered label noise problem arises whenever some elements $\yy_i$ do not match the {real} class of $\xx_i$. Several scenarii exist: in the \emph{symmetric} label noise, labels can be flipped uniformly across all the classes, whereas in the \emph{asymmetric} label noise, labels $\yy$ in the training set can be flipped with higher probability toward specific classes. We note that the first scenario, while thoroughly studied in the literature, is highly improbable in real situations: for example, it is more likely that an annotator mislabels two breeds of dogs than a dog and a car. Hence, noise in labels provided by human annotators is not symmetric since annotators  make mistakes depending on  class similarities  \cite{misra2016seeing}.

\subsection{Adversarial Regularization}
To prevent a classifier to overfit on noisy labels, we would like to regularize its decision function in areas where the local uniformity of training labels is broken. To achieve such desired local uniformity, robust optimization can be used. This amounts to enforce that predicted labels are uniform in a local neighborhood $\mathcal{U}_i$ of data point $\xx_i$. This changes the total loss function in the following way:
\begin{equation}\label{eq:robust}
\argmin_\theta \sum_{i=1}^N \max_{\xx^u_i \in \mathcal{U}_i} L(\xx_i,\yy_i,p_\theta)
\end{equation}
Because the robust optimization problem \cite{Caramanis} in~\eqref{eq:robust} is hard to solve exactly, adversarial training~\cite{Goodfellow2015,Shaham15} was proposed as a possible surrogate. Instead of solving the $\max$ inner problem, it suggests to replace it by enforcing uniformity in the direction of maximum perturbation, called the adversarial direction. Mainly used for robustness against adversarial examples, adversarial training is however not adapted to our problem, since it can enforce uniformity around a false label. Following the same reasoning, we propose to cast the problem as a regularization term of the initial loss function. The optimized loss is then
\begin{equation}
    L_{tot}(\xx_i,\yy_i,p_{\theta}
) = L_{\text{CE}}(\xx_i,\yy_i,p_{\theta}) + \beta R_{AR}(\xx_i,{p_\theta}),
\end{equation}
where $R_{AR}$ is a regularization term reading:
\begin{align}\label{AR}
    &R_{\text{AR}}(\xx_i,{p_\theta}) = D(p_{\theta}(\xx_i+ \rr_i^a), p_{\theta}(\xx_i))\;\; \nonumber \\
    &\text{with } \rr_i^a = \underset{\rr_i,\|\rr_i\| \leq \varepsilon}{\text{argmax }} D(p_{\theta}(\xx_i + \rr_i),p_{\theta}(\xx_i)).
\end{align}
\kf{Where $p_{\theta}(\xx_i)$ is the neural network prediction of the input and $p_{\theta}(\xx_i + \rr^a)$ is the neural network prediction of the adversarial input}. Basically, it minimizes an isotropic divergence $D$ between the probability output $p_{\theta}(\xx_i + \rr^a)$ and $p_{\theta}(\xx_i)$. A sound choice for $D$ can be the Kullback-Leibler (KL) divergence.
$R_{\text{AR}}$ can be seen as a (negative) measure of  local smoothness, or also as a local Lipschitz constant in the $\varepsilon$ neighborhood of $\xx_i$ with respect to the metric $D$, hence a measure of complexity of the function. $R_{AR}$ promotes local uniformity in the predictions without {using the potentially noisy label $\yy_i$}:  therefore, it reduces the influence of noisy labels, since it is {computed from the} prediction $p_{\theta}(\xx_i)$ that can be correct when the true label is not. $R_{\text{AR}}$ shares strong similarities with Virtual Adversarial Training (VAT) \cite{VATMiayto2016}, at the notable exception that we do not consider a semi-supervised learning problem and that we regularize on the labeled training {positions $\xx_i$}, where VAT is applied on unlabeled samples.

It can be shown (proof in the supplementary material 1) that this regularization  acts as a label smoothing technique:
\begin{proposition}
\label{lem1}
Let $D$ be the Kullback-Leibler divergence. Let $\gamma = \frac{\beta}{\beta+1} \in [0,1[$. Let $\yy_i^a = p_{\theta}(\xx_i+ \rr_i^a)$ be the predicted (smooth) adversarial label. Let $H$ be the entropy. The regularized learning problem $L_{tot}(\xx_i,\yy_i,p_{\theta})$ is equivalent to : $$L_{tot}(\xx_i,\yy_i,p_{\theta}) \equiv L_{\text{CE}}(\xx_i,(1-\gamma)\yy_i + \gamma \yy_i^a, p_\theta) - \gamma H( \yy_i^a).$$
\end{proposition}
 This leads to the following interpretation: instead of learning over the exact label or over a mix between the exact label and the network prediction, we learn over an interpolation between the data $\yy_i$ and the adversarial label $\yy_i^a$, while maximizing the entropy of the adversarial label (i.e. blurring the boundaries of the classifier). Related developments can be found in adversarial label smoothing (ALS)~\cite{Shafahi2018LabelSA,Goibert2019}, which aims at providing robustness against adversarial attacks. 

Yet, one of the major limit of this approach is that the regularization is conducted with the same magnitude between all classes without considering potential class similarities. 
As a consequence, a strong regularization can remove the label noise, but also hinder the ability of the classifier to separate similar classes where a complex boundary is needed. To overcome this issue, we propose to replace $D$ by a geometry-aware divergence taking into account the specific relationships between the classes for the classifier to not overfit on noisy labels when the classes are significantly different and to have complex boundaries when they are similar.{ Taking these relations into account avoid overfitting on noisy labels between dissimilar class, while allowing complex boundaries between classes with high similarity}.

\subsection{Wasserstein Adversarial Regularization (WAR)}

 \begin{figure*}[t!]
    \centering
      \includegraphics[width=\linewidth]{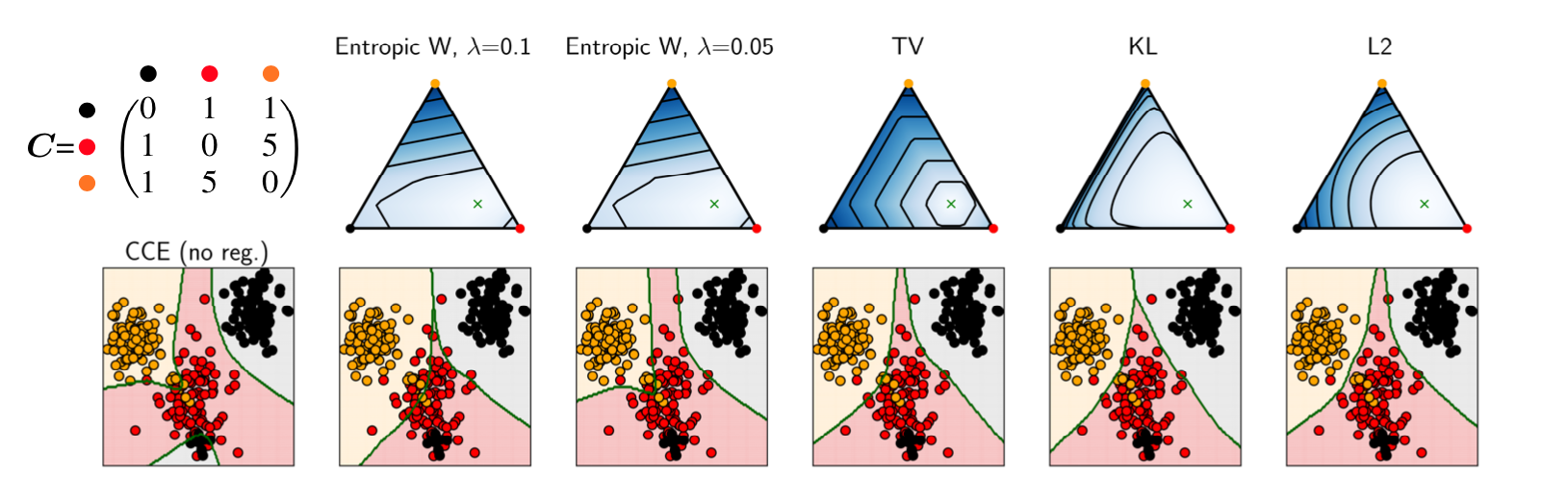}
    \caption{{Illustration of the regularization geometry for different losses in the adversarial training. (top) Regularization values on the simplex of class probabilities. Each corner stands for a class. All losses are computed with respect to a prediction represented as the green x. Colors are as follows: white is zero while darker is bigger. In the case of WAR, the ground cost $\CC$ is given on the left.  (Down) Classification boundaries when using these losses for regularization. The unregularized classifier (CCE) is given on the left.}}
    \label{fig:loss_simplex}
\end{figure*}

{To make the divergence aware of specific relationships between classes, we replace the isotropic divergence $D$ with a Wasserstein distance computed in the labels space. We name our proposed method  Wasserstein Adversarial {R}egularization.} 
 In a related way, \cite{Frogner2015} used the Wasserstein distance as a loss in a learning system between the output of the model for multi-label learning. So the OT computational complexity scales with the number of classes instead of the number of data as main OT applications. The interest of the Wasserstein distance is to take into consideration the geometry of the label space.   
We define the proposed regularization term $R_{\text{WAR}}$ as follows:
\begin{align}\label{eqn:WAR}
    &R_{\text{WAR}}(\xx_i) = OT_{\CC}^{\lambda}(p_{\theta}(\xx_i+ \rr_i^a), p_{\theta}(\xx_i))\;\; \nonumber \\
    &\text{with } \rr_i^a = \underset{\rr_i,\|\rr_i\| \leq \varepsilon}{\text{argmax }} OT_{\CC}^{\lambda}(p_{\theta}(\xx_i + \rr_i),p_{\theta}(\xx_i)).
\end{align}
$OT_{\CC}^{\lambda}$ is an optimal transport (OT) distance
\cite{COT_Peyre}. The OT problem  seeks an optimal coupling $\TT^*
\in \UU(\balpha, \bbeta) = \{\TT | \TT\geq \mathbf{0}, \TT\mathbf{1}=\balpha,  \TT^\top\mathbf{1}=\bbeta\}$
between two distributions $\balpha, \bbeta$ with respect to a ground cost $\CC \in \mathbb{R}^{n_1 \times n_2}$. $\UU(\balpha, \bbeta)$ is the space of joint probability distributions with marginals $\balpha$ and $\bbeta$. 
OT distances are classically expressed through the Wasserstein distance $
    W_{\CC}(\balpha, \bbeta) = \underset{\TT \in \UU(\balpha, \bbeta)}{\text{min }} \langle \TT, \CC \rangle$, 
where $\langle ., . \rangle$ is the Frobenius product.
Unfortunately, this distance is expensive to compute (cubical complexity). 
In practice, we will use the solution of the sharp entropic variant of the optimal transport problem \cite{CuturiSinkhorn, Luise_2018}:
\begin{align}\label{pb:variantOT}
    &OT_{\CC}^{\lambda}(\balpha, \bbeta) = \langle \TT_{\lambda}^*, \CC \rangle \;\;\nonumber \\
    &\text{with }  \TT_{\lambda}^* = \underset{\TT \in \UU(\balpha, \bbeta)}{\text{argmin }} \langle \TT, \CC \rangle - \lambda H(\TT) \nonumber
\end{align}
where $H$ denotes the entropy function and $\lambda\geq 0$ the regularization strength.
Using this regularized version has several advantages: {\em i)} it lowers the computational complexity to near-quadratic \cite{altschuler2017near}, {\em ii)} it turns the problem into a strongly convex one, for which gradients can be computed efficiently and  {\em iii)} it allows to vectorize the computation of all Wasserstein distances in a batch, which is particularly appealing for training deep neural nets. Based on \cite{Genevay_2018}, we  use the \emph{AutoDiff} framework, which approximates the derivative of this regularization with a fixed number of iterations of the Sinkhorn algorithm. {Using a large $\lambda$ results in smoothing the loss and speeds up the Sinkhorn algorithm, but drives the solution away from the true OT solution. As such, we recommend to use a small lambda to stay as close as possible to the true OT loss. }{Note that for all the experiments in the main task, we have used same regularization parameter ($\lambda=0.05$) with 20 Sinkhorn iterations, to illustrate that, while the regularization parameter is important, it is not too sensitive across datasets. To summarize, for each data in a batch, we compute the OT between the input data network prediction and the adversarial input data network prediction. Hence the OT complexity is of order $\mathcal{O}(b \times n_s \times C^2)$, where $b$ is the batch size, $n_s$ the computation budget and $C$ the number of classes. We compare the computation time of our method in the experimental section.}

\paragraph*{\bfseries Choice of ground cost.} The ground cost $\CC$ reflects the geometry of the label space. It bridges the gap between AR and WAR. An uninformative 0-1 ground cost, {\em i.e.} 0 over the diagonal and 1 everywhere else, would give the total variation (TV) loss (Remark 2.26 in \cite{COT_Peyre}), which could also be used as $D$ in
the AR framework. Below, we refer to this special case as \texttt{WAR$_{0-1}$}. To define a $\CC$ matrix encoding class relations, multiple choices are possible. Relying on expert knowledge, one could set it manually,  but this becomes unpractical when a large set of classes is present, and this knowledge is not always available, or even prone to errors. In this work, we propose two variants to estimate the ground cost: 
\begin{itemize}
    \item In absence of prior information about the nature of the source of labelling errors, we first propose to rely on semantic distances based on word embeddings such as \emph{word2vec}~\cite{Mikolov_word2vec}. Similarities between classes are then defined via Euclidean distances in the embedding space, as proposed in \cite{Frogner2015}. Finally, as our method requires large values of the cost between similar classes, we apply the function $e^{-m}$ {(where $m$ is the Euclidean distance between the two class names)} element-wise and set the diagonal of $\CC$ to 0. We denote it \texttt{WAR}$_{\text{w2v}}$; 
    \item Our second ground cost computation relies on the distance between mean of embedded data. We use a pre-trained neural network (details given in experimental section) to embed data and calculate the distance between mean of each class. {The rationale is that two close classes, in terms of their mean distances, should be harder to distinguish, and as such the method should allow for a complex boundary to discriminate them}. We denote it \texttt{WAR}$_{\text{embed}}$.
\end{itemize}
{While both options lead to crude approximations of the difficulty to discriminate two classes, we show in the experimental section that they nonetheless provide better results than the uninformative 0-1 cost and random standard normal ground cost matrices. Note that we considered pre-calculated and static ground costs. Updating dynamically the ground cost might be an interesting research direction as future work.
Other application dependent options could be designed with respect to the problem at hand. We finally note that estimating the ground cost $\CC$ could be very interesting but, despite some recent progress in metric learning for OT \cite{cuturi2014ground, huang2016supervised, Li2018}, adapting them for WAR is a difficult task, and it would be prone to overfitting due to the presence of label noise in the data.}

\paragraph*{\bfseries Function smoothness and ground metric.} Now we discuss how the proposed regularization
term regularizes the model $p_\theta$ with a smoothness
controlled by the ground metric $\CC$. It is not possible to extend the label smoothing in Proposition~\ref{lem1} to WAR because the OT distance does not admit a close form solution, but we can still show how $R_{\text{WAR}}$ promotes label smoothness. To this end, we look at the
regularization term  $OT_{\CC}^{\lambda}(\hat{p}_{\theta}(\xx), p_{\theta}(\xx +
\rr))$ for a given sample $\xx$ and a pre-computed $\rr$. We can prove (see 2 of the supplementary material) the following proposition:
\begin{proposition} Minimizing $R_{\text{WAR}}$ with a symmetric cost $\CC$ such that $C_{i,i}=0,\forall i$ is equivalent to minimizing an upper bound of a weighted total variation (TV) norm between $p_{\theta}(\xx)$ and $p_{\theta}(\xx + \rr)$. 
\begin{align}
    &\underline{c} TV(p_{\theta}(\xx),p_{\theta}(\xx +
    \rr)) \\
    &\leq \sum_k \underline{c}_k|p_{\theta}(\xx)_k- p_{\theta}(\xx +
     \rr)_k| 
    \leq OT_{\CC}^{\lambda}(p_{\theta}(\xx), p_{\theta}(\xx + \rr)) \nonumber
\end{align}\label{eq:bound_OT}
where $\underline{c}_k=\min_{i,i\neq k} c_{k,i}$ is the minimal {off-diagonal cost for} row k of $\CC$ and $\underline{c}=\min_k \underline{c}_k$ is a global minimum out of the diagonal.

\end{proposition}

By minimizing the proposed {$R_{\text{WAR}}$} regularization with $\rr$ belonging in a small ball around $\xx$,
we actually minimize a local approximation of the Lipschitz constant of
$p_\theta$. This has the effect of smoothing-out the model around $\xx$ and makes it more robust to label
noise.  One can see the effect of the cost matrix in the center term of
\eqref{eq:bound_OT}, where the values in the ground metric correspond to
a weighting of a total variation, hence controlling the effect of the
regularization {and the adversarial direction $\rr$ during adversarial computation}. Interestingly, the Wasserstein distance can be bounded both below
($\underline{c}$) and above ($\bar{c}$) by Total Variation and weighted total
variation similarly to the equation above.
Finally, in practice we minimize the expectation of the OT loss,
which means that we will penalize areas of high density similarly to a
regularization with the Sobolev {norm (i.e. penalizing the expected norm of the model gradient} \cite{mroueh2017sobolev}), while keeping a finer control of the class relations, since we use {the ground loss $\CC$ that promotes anisotropy}.

\paragraph*{\bfseries Illustration of the effect of $R_\text{WAR}$}


We illustrate AR and WAR in a simple toy 3-classes classification problem using the Scikit-learn library\cite{scikit-learn} with noise in Figure \ref{fig:loss_simplex}. 
{Each column of the figure corresponds to a divergence function $D$. The top row illustrates the values on the simplex, while the bottom row shows the classification predictions when using $D$ as adversarial regularization. From left to right, we compare the effect of training with the cross entropy alone (CCE, no regularization),  $R_\text{WAR}$ with $\lambda=0.1$ and  $\lambda=0.05$, as well as $R_\text{AR}$ with TV (that is the same as $R_\text{WAR}$ with $0-1$ loss), KL (as used in VAT \cite{VATMiayto2016}) and L2 divergences as $D$. For the classification problem, we generated two close classes (in orange and red), as well as a third (in black), which
is far from the others. Then, we introduced noisy labels (of the black class) in the region of the red class. 

On this
toy example, CCE overfits the noisy black labels, yet is able to distinguish the  red and orange classes. The $R_\text{AR}$ regularizers, being class agnostic, correct for the noisy black labels in the bottom part, but smooth the complex decision function between the orange and red classes. On the contrary, $R_\text{WAR}$ uses a different cost per pair of classes, illustrated in the top left panel of Figure~\ref{fig:loss_simplex}: the smallest cost is set between the red and black classes, which has the effect of promoting adversarial examples in that direction. This cost / smoothing relation is due to the fact that our problem is a minimization of the OT loss: in other words, the higher the cost between the classes, the less the binary decision boundary will be smoothed. Finally, the effect of the global $\lambda$ parameter can also be appreciated in the classification results: while using $\lambda=0.05$, the smoothing of the loss is decreased and the final decision boundary between the mixed classes keeps all its complexity.}


%% file: sub_part/experiments.tex

\section{Experiments}\label{sec:exp}
We evaluate the proposed approach WAR on both image classification and semantic segmentation tasks. We first showcase the performance of WAR on a series of image classification benchmarks (Section~\ref{sec:res-1}), and then consider two real world cases: first, the classification of clothing images from online shopping websites (Section~\ref{sec:res-2}) and then the semantic segmentation of land use in sub-decimeter resolution aerial images (Section~\ref{sec:res-3}). Following the community evaluation strategies, we used WAR on four different architectures depending on benchmarks. On each benchmark, WAR outperformed state of the art competitors showing the robustness of our methods on the used architectures.

\subsection{Image classification on simulated benchmark datasets}\label{sec:res-1}

\begin{table*}[!htbp]
\caption{Test accuracy ($\%$) of different models on Fashion-MNIST (F-M), Cifar-10 (C-10), and Cifar-100 (C-100) dataset with varying noise rates (0$\%$ – 40$\%$). The
mean accuracies and standard deviations averaged over the last 10 epochs of three runs are reported, and the best results are highlighted in \textbf{bold}.}
\centering
\resizebox{\textwidth}{!}{%
\begin{tabular}{lc|cccccccccccc}

\toprule
Data /& noise & CCE & Backward & Forward & Unhinge &Bootsoft& \scriptsize{CoTeaching} & \scriptsize{CoTeaching+} & D2L & SL & \texttt{Pencil} &  JoCoR & \texttt{WAR}$_{\text{w2v}}$ \\
 & &  & \cite{Patrini_2017_CVPR} & \cite{Patrini_2017_CVPR} & \cite{Rooyen_2015} &\cite{Reed2015}& \cite{Coteaching_2018} & \cite{Yu19} & \cite{D2L18} & \cite{Wang19} & \cite{pencil2019} & \cite{jocor2020} & Ours \\
\midrule
\multirow{3}{*}{\scriptsize{Fashion-Mnist}} & 0\% & 94.69$\pm$0.11 & 94.86$\pm$0.04 & 94.81$\pm$0.04 & \textbf{95.12 $\pm$ 0.03} &94.79$\pm$0.02& 94.28$\pm$0.04 & 93.62$\pm$0.01 & 94.47$\pm$0.02 
& 94.18$\pm$0.04 & 93.19 $\pm$ 0.23 &  94.60 $\pm$ 0.04 & 94.70$\pm$0.02   \\

& 20\% & 89.02$\pm$0.47 & 88.84$\pm$0.10 & 91.03$\pm$0.12 & 90.04$\pm$ 0.08 &88.17$\pm$0.11& 91.24$\pm$0.06 &92.26$\pm$0.02 & 89.12$\pm$0.15  &93.36$\pm$0.03 & 92.50 $\pm$ 0.09 & 91.01 $\pm$ 0.08 &\textbf{93.37$\pm$0.08} \\

& 40\% & 78.85$\pm$0.56 & 81.74$\pm$0.08 & 82.85 $\pm$0.2 & 78.32 $\pm$0.15 &73.84$\pm$0.28& 86.83$\pm$0.10 &86.15$\pm$0.03 &78.98$\pm$0.25 & 86.83$\pm$0.07 & 90.17 $\pm$ 0.03 & 84.86 $\pm$ 0.20 & \textbf{90.41$\pm$0.02} \\

\midrule
\multirow{3}{*}{\small{Cifar-10}} & 0\% & 91.76$\pm$0.04 & 91.63 $\pm$ 0.04   & 91.59 $\pm$ 0.03 & \textbf{92.27 $\pm$ 0.04} & 91.67 $\pm$ 0.03 & 90.12 $\pm$0.04 &88.47$\pm$0.14 & 91.29$\pm$0.02  & 90.48$\pm$0.05 & 87.90 $\pm$ 0.39 & 91.94 $\pm$ 0.02 & 91.88$\pm$0.31 \\

& 20\% & 85.26$\pm$0.09 & 84.67 $\pm$ 0.1  & 85.70 $\pm$ 0.08 & 87.09 $\pm$ 0.05 & 85.35 $\pm$ 0.8 & 86.19 $\pm$0.07 &82.97$\pm$0.25 & 86.64 $\pm$0.12 
&87.51$\pm$0.06 & 87.09 $\pm$ 0.25 & 87.93 $\pm$ 0.09 &\textbf{89.12$\pm$0.48} \\

& 40\% &  76.23$\pm$0.15 &73.49 $\pm$ 0.14 & 75.10 $\pm$ 0.15 & 77.94 $\pm$ 0.1 & 74.32 $\pm$ 0.2 & 80.87$\pm$0.09 &72.65$\pm$0.10 & 73.12 $\pm$0.43 
& 76.87$\pm$0.12 & 84.48 $\pm$ 1.04 & 82.07 $\pm$ 0.11 &\textbf{84.76$\pm$0.25} \\
\midrule

\multirow{3}{*}{\small{Cifar-100}} & 0\% & 68.60$\pm$0.09 & 69.53$\pm$0.07 & 70.12$\pm$0.07 & 70.54$\pm$0.06  & 69.81$\pm$0.04 &  65.42$\pm$0.06 &58.
93$\pm$0.14 & \textbf{70.93 $\pm$0.02}  & 68.27$\pm$0.06 & 63.32 $\pm$ 0.24 & 69.29 $\pm$ 0.26 & 68.16$\pm$0.18\\
& 20\% & 58.81$\pm$0.10 & 59.23$\pm$0.08  &  59.54$\pm$0.05 & 61.06$\pm$0.06 & 58.97$\pm$0.08 & 56.55$\pm$0.08 &44.88$\pm$0.14 & 60.90$\pm$0.03 
& 58.41$\pm$0.08 & 59.05 $\pm$ 0.34 & 60.43 $\pm$ 0.07 & \textbf{62.72$\pm$0.16} \\
& 40\% & 42.45$\pm$0.12 & 43.02$\pm$0.09 & 42.17$\pm$0.1  & 42.87$\pm$0.07 & 41.73$\pm$0.08 & 42.73$\pm$0.08 &29.94$\pm$0.34 & 42.61$\pm$0.04  & 40.97$\pm$0.12 & 45.70 $\pm$ 0.67 & 42.26 $\pm$ 0.45 &  \textbf{58.86$\pm$0.21} \\
\midrule\midrule
\multicolumn{2}{c|}{Avg,. rank} & 7.7 & 6.9 & 6.2 & 4.2 & 8.1 & 7.1 &10.1 & 6.7 &7.0 & 6.2 & 5.0 &\textbf{2.5} \\
\multicolumn{2}{c|}{noise only} & 8.8 &  8.2  & 7.1  &  5.6  &  10.0 &  5.8 &  9.5  &  7.3 & 6.0 & 3.5 & 5.0 & \textbf{1.0}\\
\bottomrule
\end{tabular}
}
\label{tab:results_classic_bench}
\end{table*}

\begin{table*}[t]
\caption{Comparison of variants of \texttt{WAR} with \texttt{AR}  with varying noise rates (0$\%$ – 40$\%$). The
mean accuracies and standard deviations averaged over the last 10 epochs of three runs are reported, and the best results are highlighted in \textbf{bold}. 
}
\centering
\resizebox{\textwidth}{!}{%
\begin{tabular}{l|ccc|ccc|ccc}

\toprule
Methods  & \multicolumn{3}{|c|}{Fashion-MNIST} & \multicolumn{3}{|c|}{CIFAR-10} & \multicolumn{3}{|c}{CIFAR-100} \\
& 0\% & 20\% & 40\% &  0\% & 20\% & 40\% & 0\% & 20\% & 40\% \\
\midrule
AR & \textbf{94.81$\pm$0.09} & 93.10$\pm$0.14 &89.74$\pm$0.10 &91.49$\pm$0.07 &88.91$\pm$0.09 & 81.98$\pm$0.25
&67.83$\pm$0.10 & \textbf{65.44$\pm$0.11} & 55.75$\pm$0.14\\
WAR$_{0-1}$& 94.60$\pm$0.03 & 90.99$\pm$0.07& 86.03$\pm$0.20
& 90.94$\pm$0.12 & 86.12$\pm$0.21 & 74.15$\pm$0.34
& 65.78$\pm$0.15 & 60.56$\pm$0.14 & 51.00$\pm$0.31 \\
WAR$_{\text{w2v}}$ &  94.70$\pm$0.02& \textbf{93.37$\pm$0.08} & \textbf{90.41$\pm$0.02}
& 91.88$\pm$0.31& {89.12$\pm$0.48}& {84.76$\pm$0.25}
& \textbf{68.16$\pm$0.18} & 62.72$\pm$0.16 & \textbf{58.86$\pm$0.21}\\
{WAR$_{\text{embed}}$} & {94.63$\pm$0.09} & {93.25$\pm$0.15} & {90.20$\pm$0.36} & {\textbf{91.88$\pm$0.12}} & {\textbf{89.93$\pm$0.02}} & {\textbf{85.08$\pm$0.32}}
& {66.58$\pm$0.21} & {62.82$\pm$0.76} & {55.79$\pm$0.25}\\
\bottomrule
\end{tabular}
}
\label{tab:results_WAR_vat}
\end{table*}

\paragraph*{{\bfseries Datasets and noisy labels simulations}}
We consider three image classification benchmark datasets: Fashion-MNIST
\cite{FMNIST}, and  CIFAR-10
/ CIFAR-100 
\cite{Kri09}. 
Fashion-MNIST consists of $60'000$ gray scale images of size $28\times 28$ with
$10$ classes. CIFAR-10 and CIFAR-100 consist of $50'000$ color images of size
$32 \times 32$ covering $10$ and $100$ classes, respectively. Each dataset also
contains $10'000$ test images with balanced classes.

Since we want to evaluate robustness to noisy labels, we simulated label noise in 
the training data only.
For all datasets, we introduced 0$\%$,
20$\%$ and 40$\%$ of noise in the labels. 
We considered only asymmetric noise, {a class-conditional label noise where each label $y_i$ in the training set is flipped into $y_j$ with probability $P_{i,j}$. {As described above, asymmetric noise is more common}  in real world scenarios than symmetric noise, where the labels are flipped uniformly over all the classes.}
For  CIFAR-10 and CIFAR-100, we follow the asymmetric noise
simulation setting by \cite{Patrini_2017_CVPR}, where class labels are
swapped only among similar classes with probability $p$ (i.e. the noise level). For Fashion-MNIST, we visually inspected the similarity between classes on a $t$-SNE plot of the activations of the
model trained on clean data; we then swapped labels between overlapping classes ($\rightarrow$: one-directional swap, $\leftrightarrow$ mutual swap): 
DRESS $\rightarrow$ T-SHIRT/TOP, COAT
$\leftrightarrow$ SHIRT, SANDAL $\rightarrow$ SNEAKER, SHIRT $\rightarrow$ PULLOVER,  ANKLE BOOT
$\rightarrow$ SNEAKER. 

\paragraph*{{\bfseries Baselines}}

We compared the proposed {WAR} with an informative $\CC$ matrix based on the \texttt{word2vec} embedding (\texttt{WAR}$_{\text{w2v}}$) against several
state-of-the-art methods: \texttt{Unhinged} \cite{Rooyen_2015},
\texttt{Bootstrapping} \cite{Reed2015}, \texttt{Forward} and \texttt{Backward}
loss correction \cite{Patrini_2017_CVPR}, \texttt{Dimensionality driven
learning  (D2L)} \cite{D2L18}, \texttt{Co-Teaching} \cite{Coteaching_2018},  \texttt{Co-Teaching+} \cite{Yu19}, \texttt{Symmetric cross entropy (SL)} \cite{Wang19}, \texttt{Pencil} \cite{pencil2019} and finally \texttt{JoCoR} \cite{jocor2020}. We then compare \texttt{WAR}$_{\text{w2v}}$ with \texttt{WAR}$_{\text{embed}}$, \texttt{AR} and \texttt{WAR$_{0-1}$}. Finally, as a baseline for all the considered methods we also included a \texttt{categorical cross entropy (CCE)} loss function.  
All the methods shared the same architecture and training procedures, as detailed in the supplementary material. 

\paragraph*{{\bfseries Model}}
Similarly to other works \cite{Coteaching_2018,VATMiayto2016},
we employed a 9-layer CNN architecture, detailed in appendix.
For   \texttt{WAR}$_{\text{w2v}}$, we set the hyper-parameters $\beta$= 10, $\lambda$= 0.05, and
 $\varepsilon$= 0.005 for all the datasets.
 The hyper-parameters of the baselines are set according to their original papers. The
noise transition matrix for the \texttt{Forward} and \texttt{Backward} method is
estimated from the model trained with cross entropy
 \cite{Patrini_2017_CVPR}.  
 The source code of WAR in PyTorch \cite{paszke2017automatic}
can be found here \footnote{\url{https://github.com/bbdamodaran/WAR}}.


\paragraph*{{\bfseries Results}}

Classification accuracies are reported in Table \ref{tab:results_classic_bench}. Results show that
 \texttt{WAR}$_{\text{w2v}}$ consistently outperforms the competitors by large margins, across noise levels and
datasets. In particular,  \texttt{WAR}$_{\text{w2v}}$ achieved improvements of 2-3\%
points on fashion-MNIST/CIFAR-10, and  15\% on CIFAR-100 at the highest
noise level. This demonstrates that the inclusion of class geometric information during training mitigates the effect of over-fitting to noisy labels. On the most noisy datasets, the most competitive method with \texttt{WAR}$_{\text{w2v}}$ is \texttt{Pencil}, however  \texttt{Pencil} tends to decrease the performance when the dataset has a smaller percentage of noisy labels which is not the case of our method. Besides  \texttt{WAR}$_{\text{w2v}}$ and \texttt{Pencil}, \texttt{JoCoR}, \texttt{Unhinged}, \texttt{Co-Teaching}, and \texttt{SL} also performed well. The \texttt{Forward} and \texttt{Backward} method performed slightly better than \texttt{CCE}, {which is most likely} due to the burden in accurate{ly estimating the} noise transition matrix. It is noted that \texttt{Co-Teaching} uses true noise estimate, and the accuracy might drop if the noise ratio is estimated directly from the noisy data. Furthermore,  performance of \texttt{Co-Teaching+}{\footnote{We used the code provided by the authors: https://github.com/xingruiyu/coteaching\_plus}} is surprising lower than the one of \texttt{Co-Teaching} on two datasets, in contrast to the observations in  \cite{Yu19}. From our experiments, we observed that \texttt{Co-Teaching+} underperforms when {the noise is class-dependent and the model considered has a wide capacity}. 


\paragraph*{{\bfseries Importance of encoding class relationships}} \label{para:class_similarities} to better assess the significance of including class relationships {and to study informative cost matrices}, we compare \texttt{WAR} in three settings: 1) \texttt{WAR$_{0-1}$}: {a version of \texttt{WAR} with an uninformative 0-1 ground cost (the cost matrix is a matrix of ones, except for the diagonal)}; {2) \texttt{WAR$_{\text{w2v}}$}: the version of WAR with the ground cost defined by \emph{word2vec} embedding (as in all other experiments in this paper); and} {3) \texttt{WAR$_{\text{embed}}$}:} a version of WAR with a ground cost defined by similarity across classes obtained by off the shelf CNNs.  
{\texttt{WAR$_{\text{embed}}$} is computed as follows: After embedding our data with a pre-trained ResNet18 \cite{resnet_2016}, we compute each class centroid (which is an imperfect statistic, as belonging to one class is determined from potentially noisy labels),  then we compute the distance between them. Finally, we apply the same exponential trick as for \texttt{WAR$_{\text{w2v}}$}. 
The intuition behind \texttt{WAR$_{\text{embed}}$} is the following: when classes are close (similar) in a relevant embedding, the noisy data will lead to closer class centroids.
Here the pretrained neural network can be seen as generic feature extractor used to assess class similarity only, similarly to \cite{Damodaran_2018_ECCV, fatras2021unbalanced, DAMODARAN_OT_noisy}. Note that for Fashion-MNIST, we computed the centroids directly the original data space.} We also compared the performance of our ground costs against 20 standard normal cost matrices drawn randomly: in all cases, our proposed ground costs were ranked at the first and second positions, 
showing the relevance to have a good crafted ground cost. Finally, we also compare these results against \texttt{AR} using KL divergence.
 We used  $\beta$=5 and $\varepsilon$ = 0.005 (similarly to \texttt{WAR} approaches), and followed the \texttt{WAR} training procedure. 

Table \ref{tab:results_WAR_vat} reports the performance of \texttt{WAR}$_{\text{w2v}}$, \texttt{WAR$_{\text{embed}}$}, \texttt{WAR$_{0-1}$} and \texttt{AR}, and shows that \texttt{WAR$_{\text{w2v}}$} is consistently
better than \texttt{AR} and \texttt{WAR$_{0-1}$} (except in  one case), and outperformed \texttt{AR} significantly by a
2-3\% margin at the highest noise level.
{When looking at the two encodings of the $\CC$ matrix, \texttt{WAR$_{\text{embed}}$} and \texttt{WAR$_{\text{w2v}}$} results are on par most of the time with the \emph{word2vec} encoding showing better performances on Fashion-MNIST and CIFAR-100 and the CNN embedding being slightly better on CIFAR-10. These results show that both embeddings are senseful and provide good priors against the label noise. We found the \emph{word2vec} embedding a good choice in our experiments, especially because they can be applied widely according to the semantics of the classes, while \texttt{WAR$_{\text{embed}}$} needs access to a meaningful pre-trained model, which can be complicated depending on the problem (e.g. for semantic segmentation or other less mainstream tasks than image classification).} Regarding computation time, all experiments were performed on a single GPU GTX TITAN. For one epoch on the CIFAR10 dataset, \texttt{WAR$_{\text{embed}}$} took 187 seconds, \texttt{AR} took 165 seconds and the usual cross-entropy loss took 57 seconds. We recall that in WAR the quadratic complexity of the Wasserstein distance\cite{altschuler2017near} only applies to the number of classes \kf{and stays of linear complexity \emph{w.r.t.} the number of samples}, hence the effect of WAR to the computational burden is relatively small. Thus the replacement of the KL divergence by the Wasserstein distance is of the same order of complexity. \kf{To demonstrate this point, we trained WAR and AR on the imagenet32 dataset with clean labels and the 9 layer CNN network. This dataset has 1000 classes and more than 1.2M $32\times32$ images. On average per epoch, the \texttt{AR} method took 1409 seconds, while \texttt{WAR} method took 1712 seconds, so  it is only 20\% more computationally expensive.}

\begin{table}[t!]
\centering
\begin{tabular}{c|cccc}
\toprule
$\beta$ & \texttt{AR} & \texttt{WAR}$_{0-1}$ & \texttt{WAR}$_{\text{w2v}}$ & \texttt{WAR}$_{\text{embed}}$ \\
\hline
0.5 & 77.49$\pm$0.18&76.80$\pm$0.22 & 77.05$\pm$0.36 & 76.96 $\pm$ 0.15\\
1 & 77.25$\pm$0.25 &76.35$\pm$0.21 & 76.90$\pm$0.37 & 77.00 $\pm$ 0.31\\
5 & 81.37$\pm$0.21 &74.76$\pm$0.15 & 80.16$\pm$0.36 &  83.13 $\pm$ 0.12\\
10 & 76.84$\pm$0.84&74.14$\pm$0.16 & 84.76$\pm$0.25 &  85.08 $\pm$ 0.32\\
20 & 57.36$\pm$0.10 & 75.58$\pm$0.18 &86.73$\pm$0.20 & 80.36 $\pm$ 0.22\\
\bottomrule

\end{tabular}
\caption{Test accuracy (in \%) of adversarial regularization methods: \texttt{AR}, \texttt{WAR}$_{0-1}$, \texttt{WAR}$_{\text{w2v}}$ and \texttt{WAR}$_{\text{embed}}$ with different $\beta$ values on CIFAR-10 dataset with 40\% noise level. The average accuracies and standard deviations over last 10 epochs are reported for one run}.
\label{table:hypparam}
\end{table}

\begin{table*}[t!]
\centering\normalsize
\caption{Test accuracy of different models on Clothing1M dataset with ResNet-50. (*) refers to results reproduced by us. ($\dagger$) means  WAR$_{\text{w2v}}$ result at the epoch showing the best validation accuracy assessed on the clean validation set.}
\begin{tabular}{c|cccccccc|cccc}
\toprule
  & \multicolumn{8}{c|}{Unsupervised} & \multicolumn{4}{c}{Using $y_{\text{val}}$}\\
  Method &CCE & CCE  & bootsoft  & D2L & GCE & SL & JoCoR & WAR$_{\text{w2v}}$ & Forward & JOF & Pencil & WAR$_{\text{w2v}}$ \\
  &\cite{Wang19} & (*) & \cite{Rooyen_2015} & \cite{D2L18} & \cite{Zhang2018} & \cite{Wang19} & (*) & Unsup.  & \cite{Patrini_2017_CVPR} & \cite{tanaka2018jof} & (*) &($\dagger$) \\
\midrule
  Acc. & 68.80 & 68.65 & 68.94  & 69.47 & 69.75  & 71.02 & 69.78 &\textbf{71.61} &  69.84 & 72.16 & 69.66 & \textbf{72.20}\\
\bottomrule
\end{tabular}
\label{tab:clothing1M}
\vspace{-4mm}
\end{table*}

{\bfseries {\bfseries Sensitivity analysis of $\beta$}:} We conducted an experimental study to analysis the sensitivity of the trade-off parameter ($\beta$) between the cross entropy and the adversarial regularization term on CIFAR-10 dataset with 40\% noise level. The experimental results with different values of $\beta$ are shown in Table \ref{table:hypparam} and the result reveals that as $\beta$ increases, \texttt{AR}, \texttt{WAR}$_{\text{embed}}$ and \texttt{WAR}$_{\text{w2v}}$ are robust to the label noise. However for the higher $\beta$, \texttt{AR} and \texttt{WAR}$_{\text{embed}}$ do over-smoothing and decreases the classification accuracy. On the other hand, \texttt{WAR}$_{\text{w2v}}$ increases the accuracy as $\beta$ increases. This behaviour shows the capability of our proposed method \texttt{WAR}$_{\text{w2v}}$ to preserve the discrimination capability between similar classes. It is noted that \texttt{WAR}$_{\text{w2v}}$ points the gradient direction towards the low cost classes, as a result it does not over smooths between the conflicting classes, thus maintaining the discrimination ability. Furthermore, \texttt{WAR}$_{0-1}$ with uninformative ground cost did not provide better results, and it is mostly similar with different values of $\beta$. This observation reinstates need of having meaningful ground cost to capture the relationship between the classes in the dataset, and to guide gradient direction with respect to the ground cost.

\subsection{Image classification with real-world label noise}\label{sec:res-2}

\paragraph*{{\bfseries Dataset}} In this section, we demonstrate the robustness of \texttt{WAR$_{\text{w2v}}$} on a large scale real world noisy label dataset, Clothing1M \cite{Xiao2015}. The Clothing1M dataset contains 1 million images of
clothing obtained from online shopping websites and has 14
classes. The labels have been obtained from text surrounding the images and are thus extremely noisy. The
overall accuracy of the labels has been estimated to $\approx$61.54\%. The dataset also contains additional manually refined clean data for training  (50k samples), validation (14k) and testing (10k). We did not use the clean training and validation data in this work, as \texttt{WAR$_{\text{w2v}}$} assumes clean labels are unavailable (which is generally a more realistic assumption). Nonetheless, we report the results obtained using the model from the epoch with the lowest validation error, calculated on the clean validation set (last column of Table~\ref{tab:clothing1M}) to show how close to it our unsupervised strategy can get and to compare with recent methods (e.g. \cite{tanaka2018jof}). Clean test data was only used to evaluate the performance of the different approaches when learning with label noise. 

\paragraph*{{\bfseries Experimental setup and Results}} Similar to \cite{Patrini_2017_CVPR,Wang19}, we used ResNet-50 pre-trained on ImageNet for a fair comparison between methods (more details in appendix). For \texttt{WAR$_{\text{w2v}}$}, the hyperparameters are similar to those of the previous experiment, except $\varepsilon=0.5$ {(details in appendix)}. For fairness of comparison, we only selected the methods { which are similar to ours (robust loss functions), and} which have the similar training methods {(optimizer, learning rate, batchsize, epochs)}  and architectures. We compared \texttt{WAR$_{\text{w2v}}$} against the competitors from \cite{Patrini_2017_CVPR,Wang19}, and also reproduced the \texttt{CCE} accuracy by our own experiments. Results are reported  in Table \ref{tab:clothing1M} and our method achieved the highest performance compared to all the baselines. {Note that Forward uses a mix of noisy and clean labels to estimate the noise transition matrix, while JOF and WAR$_{\text{w2v}}$($\dagger$) use clean validation labels only to report the test accuracy with respect to the best validation score.} 




\begin{table}[!t]\footnotesize
\caption{Per class F1 scores, average F1 score and overall accuracy ($\%$) on the test set of Vaihingen. The best results (on the noisy dataset) are highlighted in \textbf{bold}.}
\centering
\renewcommand{\tabcolsep}{5pt}
\begin{tabular}{l|c|@{\hskip3pt}c@{\hskip3pt}|@{\hskip3pt}c@{\hskip3pt}|@{\hskip3pt}c@{\hskip3pt}|@{\hskip3pt}c@{\hskip3pt}|@{\hskip3pt}c@{\hskip3pt}}
\toprule
Class & CCE & CCE & Bootsoft & \scriptsize{CoTeaching} & AR &\texttt{WAR$_{\text{w2v}}$}\\
\midrule
Training set & Clean & \multicolumn{5}{c}{Noisy} \\
\midrule
Buildings  & 90.29 & 75.06 & 88.34 & 75.1 & 81.6& \textbf{89.04}  \\
Cars   & 58.91 &14.21 & 10.98 & \textbf{26.6}& 21.6& 25.78 \\
Imperv. surf.   & 85.76 & 62.20 &\textbf{82.66} &76.9 & 70.9& 79.01 \\
Low vegetation   & 76.32 & 25.92 & 61.40 & 57.4& 57.8& \textbf{71.56} \\
Trees   & 84.72 & 70.89 & 80.49 & 77.5& 78.9& \textbf{82.92} \\
\midrule
Average F1 & 79.20 & 49.65 & 64.77 & 63.6& 62.2&\textbf{69.66}\\
\midrule
Total Accuracy & 83.89 & 63.95 & \textbf{78.87} & 74.6& 77.1 & 78.43 \\
\bottomrule
\end{tabular}
\label{tab:results_remote_sensing}

\end{table}

\subsection{Semantic segmentation of aerial images}\label{sec:res-3}

\paragraph*{{\bfseries Datasets and noisy labels simulations}}
In this experiment we consider the task of assigning every pixel of an aerial image to an urban land use category.  We use a widely used remote sensing benchmark, the ISPRS  Vaihingen semantic labeling dataset\footnote{http://www2.isprs.org/commissions/comm3/wg4/2d-sem-label-vaihingen.html}.
The data consist of 33 tiles (of varying sizes, for a total of $168'287'871$ pixels) acquired by an aircraft at the ground resolution of $9$cm. The images are true orthophotos with three spectral channels (near infrared, red, green). A digital surface model (DSM) and a normalized digital surface model (nDSM) are also available, making the input space $5$-dimensional. Among the 33 tiles, we used the initial data split (11 tiles for training, 5 for validation and 17 for testing). As ground truth,  six land cover classes (impervious surfaces, building, low vegetation, tree, car, background/clutter) are densely annotated.

 \begin{figure}[t!]
     \centering
     \includegraphics[width=.9\columnwidth]{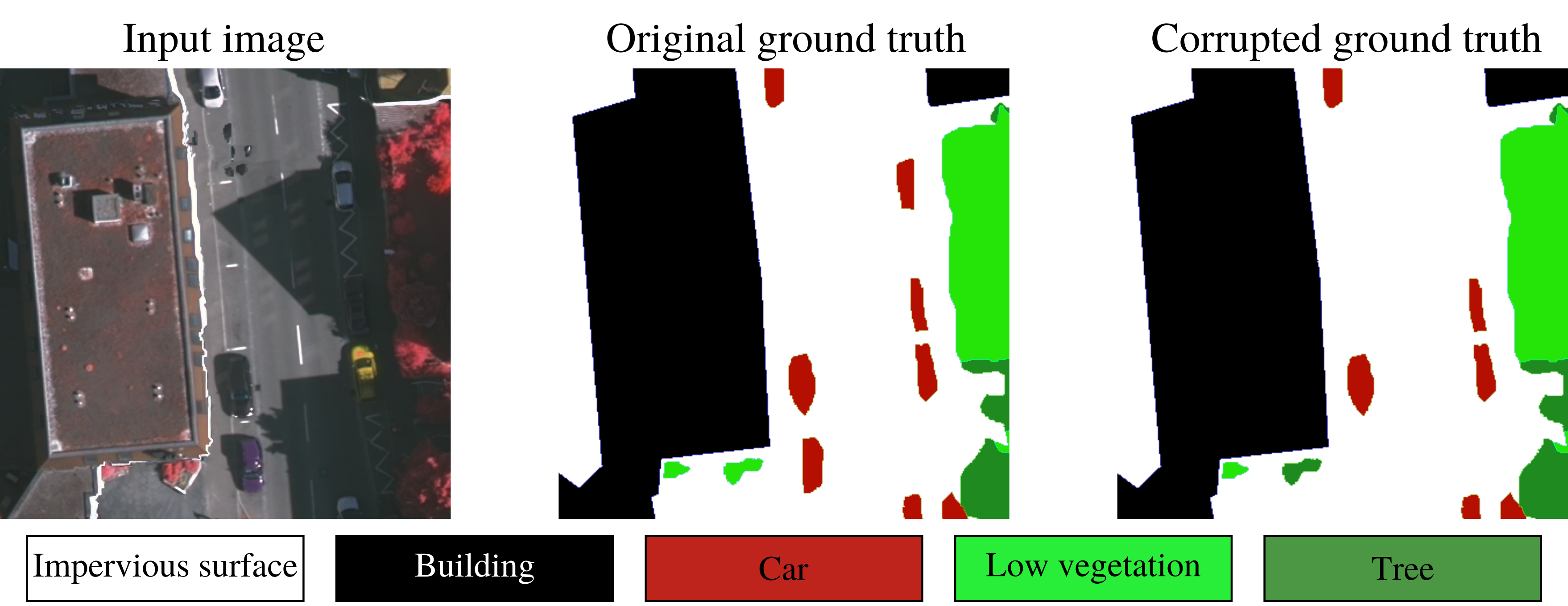}
     \caption{\label{fig:compGT}Comparison of the original and the corrupted ground truths for the semantic segmentation experiment.}
 \end{figure}

\begin{figure*}[t!]\footnotesize
    \centering
    \includegraphics[width=\textwidth]{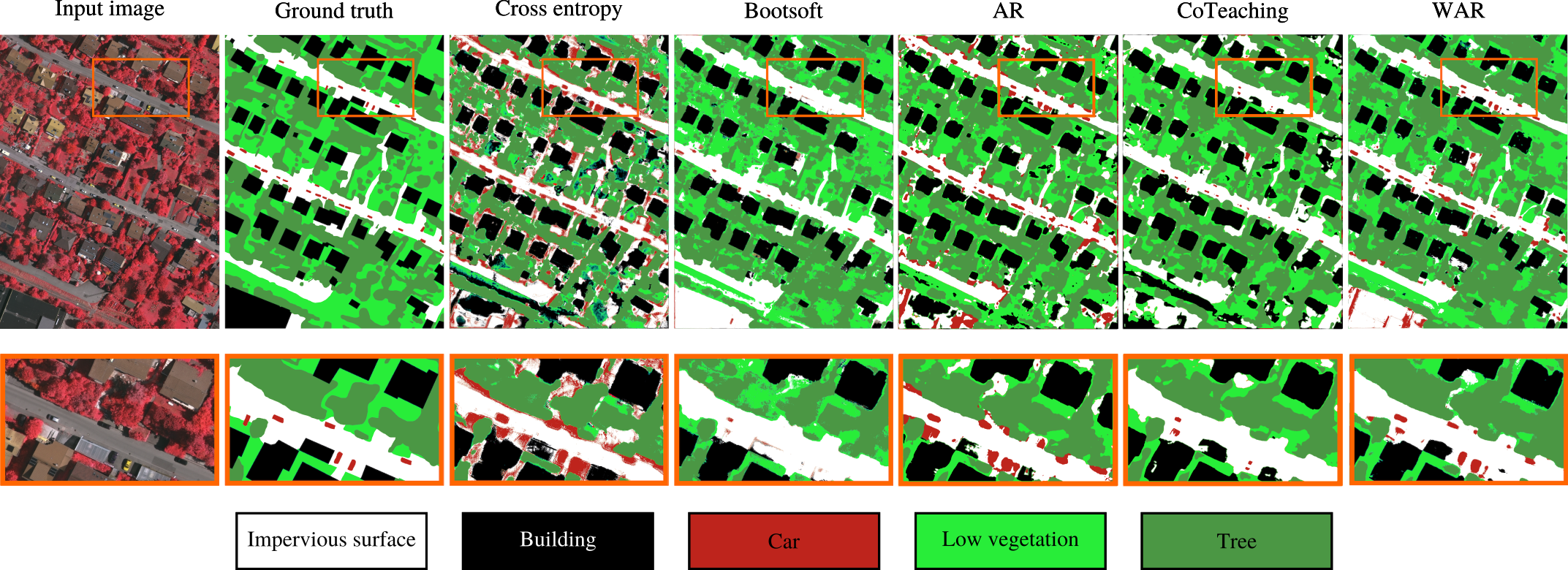}
    \caption{\label{fig:resVaihingen}Semantic segmentation maps obtained on the test set of the ISPRS Vaihingen dataset (tile \#12 of the original data). The top row shows the full image, and the second row shows a close-up of the area delineated in orange.}
\end{figure*}

\begin{table*}[t!]
    \caption{{Test accuracy on CIFAR10 dataset with 40\% openset samples from SVHN and ImageNet32. }}
    \centering
    \begin{tabular}{l|cccccc}
    \hline
    Dataset & CCE & Forward & Iterative & Co-teaching & Co-teaching+ &  \texttt{WAR}$_{\text{w2v}}$ \\
    \hline
    CIFAR-10+ 40\% SVHN & 67.45   & 56.70 & 77.73 &80.95 & 77.53 & \textbf{82.03}  \\
    CIFAR-10+ 40\% ImageNet32 & 66.69 & 66.77 & 79.38 & 80.34 & 75.47 & \textbf{80.61}\\
    \hline 
    \end{tabular}
    \label{tab:results_openset}
\end{table*}

We simulated label noise by swapping labels at the object level rather than flipping single pixels. An object is the connected component of pixels sharing the same label. We also focused on plausible labeling errors: for
instance, a car could be mislabeled to an impervious surface, but not to a
building or a tree. Following this methodology, a third of the connected
components had the label flipped. An example of the corrupted label data is shown in Figure \ref{fig:compGT}.

\paragraph*{{\bfseries Model}}
We used a U-Net architecture \cite{ronneberger2015u}, modified to take the 5 channels input data as inputs. More details about the training procedure are reported in the supplementary material. This architecture has been shown to perform well on the semantic segmentation task in general, and on this dataset in particular \cite{xu2018building}.
Using this methodology, we obtain an overall accuracy on the clean data of 83.89\%, which is close to the state of the art for this dataset.

\paragraph*{{\bfseries Results}}
We compare \texttt{WAR$_{\text{w2v}}$} with standard \texttt{CCE}, \texttt{Bootsoft}, \texttt{Co-Teaching} and \texttt{AR}. The
results, computed on the full test ground truth (including boundaries) and averaged over 2 runs, are reported in
Table \ref{tab:results_remote_sensing}. Note that the classes are unbalanced and, for most of them, the F1-score is
improved using \texttt{WAR$_{\text{w2v}}$}, except for the dominant class (impervious surfaces). This
leads to a much higher average F1-score using \texttt{WAR$_{\text{w2v}}$} (compared to its competitors), while the overall accuracy is
only slightly decreased compared to \texttt{Bootsoft}. This behavior can be seen in the maps
shown in Figure \ref{fig:resVaihingen}. We
can see in the close-ups that \texttt{Bootsoft} performs poorly in detecting the cars,
which are often confused with generic impervious surfaces.

\subsection{{Open Set Noisy Labels}}\label{exp:sec-4}

{As a final benchmark, following \cite{Wang18,Yu19} we evaluate our proposed method on  realistic open set noisy labels scenarios. The open set noisy datasets are created by replacing some training images with out of domain images, while keeping the labels and number of images per class unchanged. In our experiments, we simulated open set noisy datasets following \cite{Wang18,Yu19} for CIFAR 10 dataset by replacing images with others coming  either from the SVHN or ImageNet32 (images of size 32$\times$32) \cite{Aaron16} datasets.}

\paragraph*{{\bfseries Experimental set-up and Results.}}
We used the architecture (6 convolutional layers and one fully connected layer) and experimental setting as in \cite{Wang18} to ensure a fair comparison with the state-of-the-art methods. Additional details about model architecture and training procedure are described in the supplementary material. Table \ref{tab:results_openset} reports the classification accuracy on the CIFAR-10 dataset with 40\% of openset noise. The scores for the Iterative and Forward method are from \cite{Wang18}, while for other competing methods we report our replicated scores. Our method outperformed the competitors on both openset noisy datasets. Our method is significantly better than all competitors for openset samples from SVHN, while for ImageNet32 openset samples we are slightly better than Co-teaching, and significantly better than remaining methods. Note that our method does not need to reject openset samples on contrary to competitors.


%% file: sub_part/conclusion.tex
\section{Conclusion and discussion}

In this paper, we proposed Wasserstein Adversarial Regularization (WAR) to address the problem of learning with noisy labels. Using a ground cost based on class similarites or prior knowledge, we are able to change the geometry of the regularization loss according to class
similarities. 
We compare WAR with state of the art algorithms on the Fashion-MNIST, CIFAR-10, CIFAR-100 benchmarks with noisy labels up to 40\%. WAR outperformed  state of the art results on all benchmarks. Furthermore, we proved that WAR performs accurately on real life problems in both classification and semantic segmentation problems.

Our regularization scheme is versatile, and could be used in conjunction with other methods involving samples rejection. Future works will also consider exploring other strategies to define the ground cost, beyond the current {\em a priori} setting: such cost could be for instance derived from auxiliary tasks that do not involve labels, or by inspecting the confusion matrix score, obtained on the noisy validation set, along the learning process. \kf{A dynamic update of the class matrix is also a relevant research avenue to be studied in the case the noise statistics of the data change over time.} Moreover, even if in this paper we focused on the label noise problem, WAR remains a generic regularization scheme that could be applied to other classical learning problems as
enforcing adversarial robustness or semi-supervised learning.
In all those
cases, disposing of a loss that is sensitive to inter classes relationships might
help in designing more robust models that can focus on specific class imbalance or
permutations.

%% file: sub_part/appendix.tex
\setcounter{section}{0}
\renewcommand\thesection{\Alph{section}}


\subsection{Links between \texttt{AR} and adversarial label smoothing}\label{suppl:KL}
This part gathers the proof for Proposition 1. The total learning loss for one sample $(\xx, \yy)$ is:
\begin{align}
    L_{\text{tot}}(\xx, \yy, p_\theta) = L_{\text{CE}}(\xx,\yy, p_\theta) + \beta R_{\texttt{AR}}(\xx, p_\theta) 
\end{align}
where $L_{\text{CE}}$ is the cross entropy loss:
\begin{align}
   L_{\text{CE}}(\xx,\yy, p_\theta) = - \sum_c \yy^{(c)} \log  p_\theta(\xx)^{(c)}
\end{align}
We write the $R_{\texttt{AR}}$ regularization as:
\begin{align}\label{VAT}
    R_{\texttt{AR}}(\xx, p_\theta) &= D_{KL}( p_{\theta}(\xx + \rr^a), p_{\theta}(\xx))\nonumber\\
    \text{where } \rr^a &= \underset{\rr,\|\rr\| \leq \varepsilon}{\text{argmax }} D_{KL}(p_{\theta}(\xx + \rr), p_{\theta}(\xx)).
\end{align}
We have that:
\begin{align}
&D_{KL}(p_{\theta}(\xx + \rr^a),p_{\theta}(\xx)) \nonumber\\
&\qquad =  \sum_c p_{\theta}(\xx+ \rr^a)^{(c)} \log \frac{p_{\theta}(\xx+ \rr^a)^{(c)}}{p_{\theta}(\xx)^{(c)}}\nonumber\\
&\qquad =  \sum_c p_{\theta}(\xx+ \rr^a)^{(c)} \log p_{\theta}(\xx+ \rr^a)^{(c)} \nonumber \\
& \qquad \qquad - \sum_c p_{\theta}(\xx+ \rr^a)^{(c)} \log p_{\theta}(\xx)^{(c)} \nonumber\\
&\qquad =   - \sum_c p_{\theta}(\xx+ \rr^a)^{(c)} \log p_{\theta}(\xx)^{(c)} - H( p_{\theta}(\xx+ \rr^a)) .
\end{align}
where $H$ is the entropy function. Consequently, the total loss can be rewritten as:
\begin{align}\label{global}
    L_{\text{tot}}(\xx, \yy, p_\theta) =& - \sum_c (\yy^{(c)} + \beta p_{\theta}(\xx+ \rr^a)^{(c)}) \log  p_\theta(\xx)^{(c)} \nonumber \\
    &- \beta H( p_{\theta}(\xx+ \rr^a)) 
\end{align}
Here $\beta \in \mathbb{R}^+$. Let $\beta = \frac{\epsilon}{1-\epsilon}$ with $\epsilon \in [0,1[$. We have the following equivalence:
\begin{align}\label{global2}
     &(1-\epsilon) L_{\text{tot}}(\xx, \yy, p_\theta) \nonumber \\
     &  = - \sum_c ( (1-\epsilon)\yy^{(c)} + \epsilon p_{\theta}(\xx+ \rr^a)^{(c)}) \log  p_\theta(\xx)^{(c)} \nonumber \\
     & \quad - \epsilon H( p_{\theta}(\xx+ \rr^a))\nonumber\\
     & = L_{\text{CE}}(\xx,\underbrace{(1-\epsilon)\yy + \epsilon p_{\theta}(\xx+ \rr^a))}_\text{Interpolated label}, p_\theta) - \epsilon \underbrace{H( p_{\theta}(\xx+ \rr^a))}_\text{adversarial label entropy}
\end{align}
{\em i.e.} That's why learning with the total loss is equivalent to learn on an interpolated label and maximizing the entropy of adversarial labels.

\subsection{Links between \texttt{WAR} and label smoothing}
\label{ssec:app_lemma2}
In this subsection we aim to prove the following relations
\begin{equation*}
    \underline{c} TV(\balpha,\bbeta)
    \leq \sum_i \underline{c}_i|\alpha_i-\beta_i| 
    \leq OT_{\CC}^{\lambda}(\balpha,\bbeta)
    \label{eq:bound_OT_app}
\end{equation*}

First we recall the definition of the Wasserstein distance
\begin{equation}
     W_{\CC}(\balpha, \bbeta) = \underset{\TT \in \UU(\balpha, \bbeta)}{\text{min }} \langle \TT, \CC \rangle
     \label{eq:wass}
\end{equation}
where $\UU(\balpha, \bbeta)=\{\TT | \TT\geq \mathbf{0}, \TT\mathbf{1}=\balpha,  \TT^\top\mathbf{1}=\bbeta\}$. It is well known and obvious from \cite{CuturiSinkhorn,Luise_2018} that the optimal OT matrix of regularized OT $\TT_\lambda^\star$ leads to a larger OT loss than the exact OT solution of the problem above $\TT^\star$. This means that
\begin{equation}
      W_{\CC}(\balpha, \bbeta) =  \langle \TT^\star, \CC \rangle\leq \langle \TT^\star_\lambda, \CC \rangle= OT_{\CC}^{\lambda}(\balpha,\bbeta)\label{eq:proof1}
\end{equation}
and the relation is strict when $\lambda>0$. 

Now if we suppose that the cost matrix is symmetric and $C_{i,i}=0$ and $C_{i,j}>0$ when $i\neq j$ then solving \eqref{eq:wass} means that the maximum amount of mass on the diagonal of $\TT^\star$ since it leads to a $0$ cost. Under constraints $\UU(\balpha, \bbeta)$ this maximum amount is equal to $\TT^\star_{i,i}=\min(\alpha_i,\beta_i),\forall i$. 
This implies that for a given row $i$ in $\TT^\star$ the amount of mass not on the diagonal row $i$ is $\sum_{j\neq i}\TT^\star_{i,j}=\max
 (\alpha_i-\beta_i,0) $ because of the left marginal constraint in $U$. Note that a similar result can be expressed with the column j such that the mass not on the diagonal of column $j$ is
 $\sum_{i\neq j}\TT^\star_{i,j}=\max (\beta_j-\alpha_j,0)$. This obviously means that for a given column/row index $k$ we have $\sum_{i\neq k}\TT^\star_{i,k}+\TT^\star_{k,i}=|\alpha_k-\beta_k|$.

Let's write $A_k=\sum_{i\geq k,j\geq k}T^\star_{i,j} C_{i,j}$. We have that $W_{\CC}(\balpha, \bbeta)=A_1$. Now we remark that $$A_k - A_{k+1}\geq \underline{c}_k |\alpha_k-\beta_k|.$$ We can write that $$A_1=A_1 - \sum_{k=2} A_k + \sum_{k=2} A_k.$$ Since $A_N=0$ because $C_{N,N}=0$, it turns out that $A_1=\sum_{k=1} (A_k-A_{k+1})$. Lower bounding every elements of the sum by the previous minoration gives that:
\begin{align}
    A_1&=\sum_{k=1} (A_k-A_{k+1})\\
    &\geq\sum_{k=1} \underline{c}_k |\alpha_k-\beta_k|\\
    &\geq\underline{c}TV(\balpha,\bbeta)
\end{align} 
which gives the desired results.
\subsection{Adversarial samples computation for \texttt{WAR} }
\texttt{WAR} requires an efficient
computation of  adversarial samples. Following ~\cite{VATMiayto2016}, we choose the following computation model. 
One could use the gradient with respect to the input $\rr$ but because of differentiability, it vanishes in $\rr=0$. When we approximate $OT_{\CC}^{\lambda}$ in
$\rr=0$ through the second order Taylor expansion, we have
\begin{equation}
   OT_{\CC}^{\lambda}(p_{\theta}(\xx), p_{\theta}(\xx + \rr)) \underset{r=0}{\sim} \frac{1}{2} \rr^t \HH_{\rr} \rr.
\end{equation}
 However, computing the hessian $\HH_{\rr}$ with respect to $\rr=0$ is
 costly. Instead we use the power iteration method \cite{Golub_2000} to estimate the dominant hessian's
 eigenvector that represent the direction in which  the classification function will change the most. The algorithm is repeated $k_{\text{max}}$ times, but both the literature and our results suggest that only one iteration is sufficient to achieve state of the art results. Once the adversarial direction $\dd$ is defined, one can obtain the adversarial example
with $\rr=\varepsilon\dd/\|\dd\|_2$, by projecting onto the ball of radius $\varepsilon$.

\subsection{Model architecture, implementation details and training procedure} \label{app:exp_details}
\subsubsection{Benchmark datasets}
We have used a 9 layer CNN following \cite{Coteaching_2018} for the three image classification benchmark datasets: Fashion-MNIST, CIFAR10, and CIFAR100 as shown in Table \ref{table:Architecture}. Between each layer we use a batch norm
layer, a drop-out layer and a leaky-relu activation function with slope of
$0.01$. We use the Adam optimizer for all our networks with an initial learning
rate of 0.001 with coefficient $(\beta_1, \beta_2) = (0.9, 0.999)$ and with
mini-batch size of 256. 
 The learning rate is divided by 10 after epochs 20 and 40 for Fashion-MNIST (60
 epochs in total), after epochs 40 and 80 for CIFAR-10 (120 epochs in total),
 and after epochs 80 and 120 for CIFAR-100 (150 epochs in total). While training \texttt{WAR}, we set
 $\beta$ = 0 for 15 epochs for faster convergence, as we observed that the network does not overfit
 on noisy labels at early stages of training. 
The input images are scaled between [-1, 1] for Fashion-MNIST, and  mean subtracted  for the CIFAR10, and CIFAR100 datasets before feeding into the network. The proposed method \texttt{WAR}, \texttt{AR}, and cross entropy loss functions are implemented in PyTorch, and for the
JoCoR{\footnote{https://github.com/hongxin001/JoCoR}}, \texttt{Pencil}{\footnote{https://github.com/yikun2019/PENCIL}}, Co-teaching{\footnote{https://github.com/bhanML/Co-teaching}}, Co-teaching+{\footnote{https://github.com/xingruiyu/coteaching\_plus}}, SL{\footnote{https://github.com/YisenWang/symmetric\_cross\_entropy\_for\_noisy\_labels}}  method we used the PyTorch code provided by the authors. For the rest of the state-of-the-art methods (dimensionality driven learning{\footnote{https://github.com/xingjunm/dimensionality-driven-learning}}, forward, backward loss correction and robust loss functions {\footnote{https://github.com/giorgiop/loss-correction}}: unhinged and boot strapping) the experiments are conducted using the Keras code provided by respective authors. We used similar layer initialization for all the methods in Pytorch and Keras.
\begin{table}[h!]
\centering
\begin{tabular}{c|c|c}
 \hline
    Fashion-MNIST & CIFAR-10 & CIFAR-100 \\
    \hline
 28$\times$28$\times$1  &   32$\times$32$\times$3 & 32$\times$32 $\times$3\\
        \hline
            \multicolumn{3}{c}{3$\times$3 conv, 128 LReLU }\\
            \multicolumn{3}{c}{3$\times$3 conv, 128 LReLU }\\
            \multicolumn{3}{c}{3$\times$3 conv, 128 LReLU }\\
            \hline
            \multicolumn{3}{c}{2$\times$2 max-pool, stride 2}\\
            \multicolumn{3}{c}{dropout, p=0.25}\\
            \hline
            \multicolumn{3}{c}{3$\times$3 conv, 256 LReLU }\\
            \multicolumn{3}{c}{3$\times$3 conv, 256 LReLU }\\
            \multicolumn{3}{c}{3$\times$3 conv, 256 LReLU }\\
            \hline
            \multicolumn{3}{c}{2$\times$2 max-pool, stride 2}\\
            \multicolumn{3}{c}{dropout, p=0.25}\\
            \hline
            \multicolumn{3}{c}{3$\times$3 conv, 512 LReLU }\\
            \multicolumn{3}{c}{3$\times$3 conv, 256 LReLU }\\
            \multicolumn{3}{c}{3$\times$3 conv, 128 LReLU }\\
            \hline
            \multicolumn{3}{c}{avg-pool}\\
            \hline
 dense 128 $\rightarrow$ 10 &   dense 128 $\rightarrow$ 10 & dense 128 $\rightarrow$ 100\\
 \hline
\end{tabular}
\caption{CNN models used in our experiments on Fashion-MNIST, CIFAR-10 and CIFAR-100.}
\label{table:Architecture}
\end{table}

\paragraph*{{\bfseries Discussion on Co-Teaching+}}
In order to analysis low performance of Co-Teaching+\cite{Yu19}, we conducted a series of experiments on \kf{}{the benchmark} CIFAR-10 with 40\% of noisy label. \kf{}{We consider two neural networks, a 2-layer CNN and a 9-layer CNN, on two noisy label settings, pairflip noisy labels and class dependant noisy labels. First, for a similar 2-layer CNN architecture} as the one used in their paper \kf{}{and for a pairflip noisy label setting, we reproduced their results} (39\%, 43\% for Co-Teaching and Co-Teaching+ at 120 epochs). Next we conducted experiments with the \kf{}{different noisy label settings and architectures to analysis sensitiveness of Co-Teaching+}, and the results are shown in Table \ref{tab:coteach+}. \kf{}{To summary, for a pairflip noise and the 2-layer CNN neural network}, Co-Teaching+ performs better than Co-Teaching, however it is the opposite when \kf{}{the class dependant noise setting} is considered. When the \kf{}{9-layer CNN} model is considered, Co-Teaching+ is inferior in both noise settings.

To understand this behaviour, we observed the pure ratio of the selected instances of both methods, and found that the pure ratio of selected instances decreases after few epochs of Co-Teaching+ loss (Co-Teaching+ uses a warm-up strategy with Co-Teaching method for 20 epochs).  For example with CIFAR-10 at 40\% noise, the pure ratio of selected instances is approximately around 57\%, whereas in Co-Teaching it is around 70\%. It is noted that \cite{Yu19} considered an unrealistic noise simulation (Pairflip), which flips the class labels with respect to the successive classes without considering class similarities, where as in our paper we have considered a class dependent noise, which flips the label according to the class \kf{}{similarities \cite{Patrini_2017_CVPR}, explaining the perfomances we reported in our paper.}

\begin{table}[t!]
\centering
\caption{Test accuracy of Co-Teaching and Co-Teaching+ with different model architectures and noise settings on CIFAR-10 with 40\% noise level.}
\begin{tabular}{l|ccc}
\toprule
Model & Noise & Co-Teaching & Co-Teaching+ \\
\midrule
2 layer CNN & pairflip & 47\% & 52\% \\
 & class dependent & 56\% & 54\% \\
 \midrule
 9 layer CNN & pairflip & 78\% & 64 \% \\
  & class dependent & 80\% & 72\% \\
  \bottomrule
\end{tabular}
\label{tab:coteach+}
\end{table}

\subsubsection{Clothing 1M dataset}

Regarding the training procedure for the Clothing1M dataset we give the following details. Data pre-processing includes resizing the image to 256 x 256, center cropping a 224 x 224 patch from the resized image, and performing mean subtraction. We used a batch size of 64 and learning rate of 0.002 to update the network with Stochastic Gradient Descent (SGD) optimizer with a momentum of 0.9, and weight decay of 0.001. The learning rate is divided by 10 after 5 epochs (10 epochs in total). We set $\varepsilon=0.5$ and divided by 10 (after 5 epochs) when the training loss is not decreasing. For \cite{pencil2019}, the first stage was for three epochs with step size of $2e^{-3}$, the second for 6 epochs with step size $2e^{-4}$, and the last stage was for 6 epochs with step size $2e^{-5}$. For JoCoR \cite{jocor2020}, we got better results by setting the learning rates as follows: we use the step size $2e^{-3}$ for the three first epochs, then we divided it by 10 for 6 epochs and finally used $2e^{-5}$ for the remaining 6 epochs.

\subsubsection{Vaihingen dataset}
We now give the training procedure for our U-net on the ISPRS  Vaihingen semantic labeling dataset. The network was trained for 300 epochs (with $90^\circ$, $180^\circ$ or $270^\circ$ rotations and vertical or horizontal flips as data augmentation) using the Adam optimizer with an initial learning rate of $10^{-4}$ and coefficients $(\beta_1, \beta_2) = (0.9, 0.999)$. After 10 epochs, the learning rate is set to $10^{-5}$. Furthermore, we predict on the full image using overlapping patches (200 pixels overlap) averaged according to a Gaussian kernel centered in the middle of the patch ($\sigma$ = 1).

\subsubsection{Openset Noisy Labels}

{For the openset noise experiments, we used a 6 layers CNN with one fully connected layer following \cite{Wang18,Yu19} (see Table \ref{table:openset_architecture}). Between each convolutional layer we used a batch normalization layer with momentum of 0.1. The images are scaled in the range  [-1, 1] and the networks are trained by SGD with learning rate 0.01,
weight decay 10$^{-4}$ and momentum 0.9. The learning rate is divided by 10 after 40 and 80 epochs (100 in total).  
}

\begin{table}[t]
\centering
\begin{tabular}{c|c}
 \hline
     CIFAR-10+SVHN & CIFAR-10+ImageNet32 \\
    \hline
   32$\times$32$\times$3 & 32$\times$32 $\times$3\\
        \hline
            \multicolumn{2}{c}{3$\times$3 conv, 64 ReLU }\\
            \multicolumn{2}{c}{3$\times$3 conv, 64 ReLU }\\
            \multicolumn{2}{c}{2$\times$2 max-pool, stride 2}\\
            \hline
            \multicolumn{2}{c}{3$\times$3 conv, 128 ReLU }\\
            \multicolumn{2}{c}{3$\times$3 conv, 128 ReLU }\\
            \multicolumn{2}{c}{2$\times$2 max-pool, stride 2}\\
            \hline
            \multicolumn{2}{c}{3$\times$3 conv, 196 ReLU }\\
            \multicolumn{2}{c}{3$\times$3 conv, 16 ReLU }\\
            \multicolumn{2}{c}{2$\times$2 max-pool, stride 2}\\
            \hline
   \multicolumn{2}{c} {dense 256 $\rightarrow$ 10} \\
 \hline
\end{tabular}
\caption{CNN models used in our open set noisy label experiments on CIFAR-10 with openset noise from SVHN and ImageNet32.}
\label{table:openset_architecture}
\end{table}